\documentclass[letterpaper]{article} 
\usepackage[preprint]{aaai2027}  
\usepackage{times}  
\usepackage{helvet}  
\usepackage{courier}  
\usepackage[hyphens]{url}  
\usepackage{graphicx} 
\usepackage{natbib}  
\usepackage{caption} 
\usepackage{subfig}
\usepackage{algorithm}
\usepackage{algorithmic}
\usepackage{newtxmath}
\usepackage{mathtools}
\usepackage{multirow}
\usepackage{array}
\usepackage{booktabs}
\usepackage{newfloat}
\usepackage{listings}
\DeclareCaptionStyle{ruled}{labelfont=normalfont,labelsep=colon,strut=off} 
\floatstyle{ruled}
\newfloat{listing}{tb}{lst}{}
\floatname{listing}{Listing}
\usepackage{xcolor}%

\usepackage{booktabs}
\usepackage{multirow}
\usepackage{graphicx}
\usepackage[table]{xcolor}
\usepackage{tabularx}
\usepackage{array}

\title{Free-Flow Class-Incremental Learning: Towards Robust CIL \\ under Variable Class Arrivals}
\author{
    Zhiming Xu\textsuperscript{\rm 1,2},
    Baile Xu\textsuperscript{\rm 1,2},
    Furao Shen\textsuperscript{\rm 1,2,$\dagger$},
    Jian Zhao\textsuperscript{\rm 4},
    Suorong Yang\textsuperscript{\rm 1,3}
}
\affiliations{}

\usepackage{bibentry}

\makeatletter
\newcommand\blfootnote[1]{%
  \begingroup
  \renewcommand\thefootnote{}%
  \footnotetext{#1}%
  \addtocounter{footnote}{-1}%
  \endgroup
}
\makeatother

\begin{document}

\maketitle

\blfootnote{%
\textsuperscript{\rm 1}National Key Laboratory for Novel Software Technology, Nanjing University, Nanjing, China. 
\textsuperscript{\rm 2}School of Artificial Intelligence, Nanjing University, Nanjing, China. 
\textsuperscript{\rm 3}School of Computer Science, Nanjing University, Nanjing, China. 
\textsuperscript{\rm 4}School of Electronic Science and Engineering, Nanjing University, China.
$\dagger$ Correspondence to: Furao Shen $<$frshen@nju.edu.cn$>$.
}

\begin{abstract}
Class-incremental learning (CIL) is commonly evaluated under predefined schedules with fixed or nearly equal class increments, leaving irregular class-arrival scenarios underexplored.
However, practical CIL systems may need to update whenever new categories emerge, without forcing them into balanced task partitions.
We formalize this setting as Free-Flow Class-Incremental Learning (FFCIL), where the number of newly arriving classes can vary substantially across learning stages.
We show that variable class arrivals alter the class composition of incremental training, the reliability of new-class classifier statistics, and the consistency of representations learned across increments, causing clear performance degradation in both conventional and pre-trained model (PTM)-based CIL methods.
To improve robustness under FFCIL, we introduce a general framework consisting of Class-Wise Mean (CWM), which replaces instance-wise loss aggregation with class-wise averaging; Dynamic Intervention Weight Alignment (DIWA), which adjusts new-class weight calibration according to the current increment size; and Head-Agnostic Alignment (HA), which performs feature-level correction for PTM-based methods using current and previous-class feature supervision.
Extensive experiments across diverse methods, datasets, and class-arrival schedules demonstrate the general impact of FFCIL and the consistent effectiveness of our framework.

\end{abstract}
\section{Introduction}
Over the past decade, deep networks have achieved remarkable success across diverse applications~\cite{ye2019learning,chen2022learning,yang2024entaugment}. However, the underlying data distribution and category set are presumed to remain static post-training.
This simplifying premise catastrophically fails in real-world scenarios, where models inevitably encounter non-stationary data streams and the sequential emergence of novel classes~\cite{gomes2017survey}.
Because standard training under such dynamic conditions leads to severe catastrophic forgetting, CIL ~\cite{zhou2024class,wang2022beef,liangloss,xu2025dual}  has emerged as a vital paradigm. It continuously learns new concepts from non-stationary streams while preserving the integrity of historical knowledge.

Existing CIL paradigms can be broadly categorized into traditional methods without Pre-Trained Models (PTMs) and PTM-based methods. Traditional CIL methods include replay-based methods~\cite{rolnick2019experience,wang2025enhancing} that store or generate representative samples, regularization-based and distillation-based methods~\cite{nguyen2018variational,wu2019large,bian2024make} that penalize parameter shifts or preserve functional consistency, and expansion-based methods~\cite{yan2021dynamically,zhou2022model,zheng2025task} that accommodate new classes by expanding the feature extractor across tasks. In contrast, PTM-based methods \cite{wang2025tuna,he2025cllora} typically freeze a powerful pretrained backbone and perform continual adaptation by learning lightweight tuning modules or prediction heads.
Despite the progress of both paradigms, their evaluation protocols usually inherit a strong scheduling assumption: the label space is divided into fixed or nearly equal class increments. For instance, a 100-class benchmark is often split into ten learning stages with ten new classes per stage. This convention facilitates clean comparison, but it also removes an important source of difficulty from continual learning: the number of new classes arriving at each update may itself be highly irregular.

\begin{figure}[!t]
  \centering
  \subfloat[FFCIL case]{
    \includegraphics[height=1.45in,keepaspectratio]{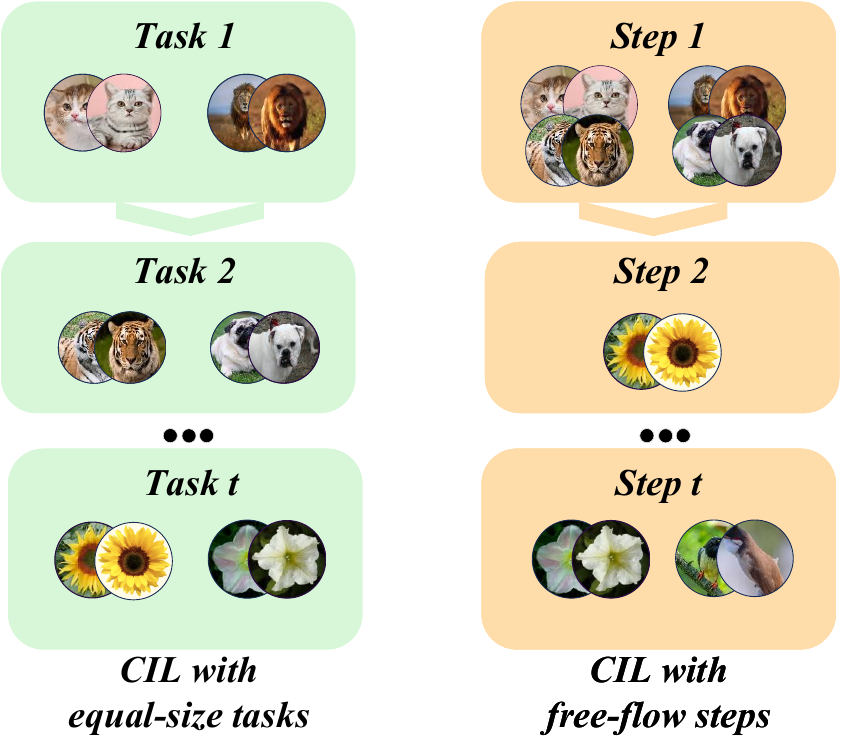}}
  \subfloat[Performance]{
    \includegraphics[height=1.45in,keepaspectratio]{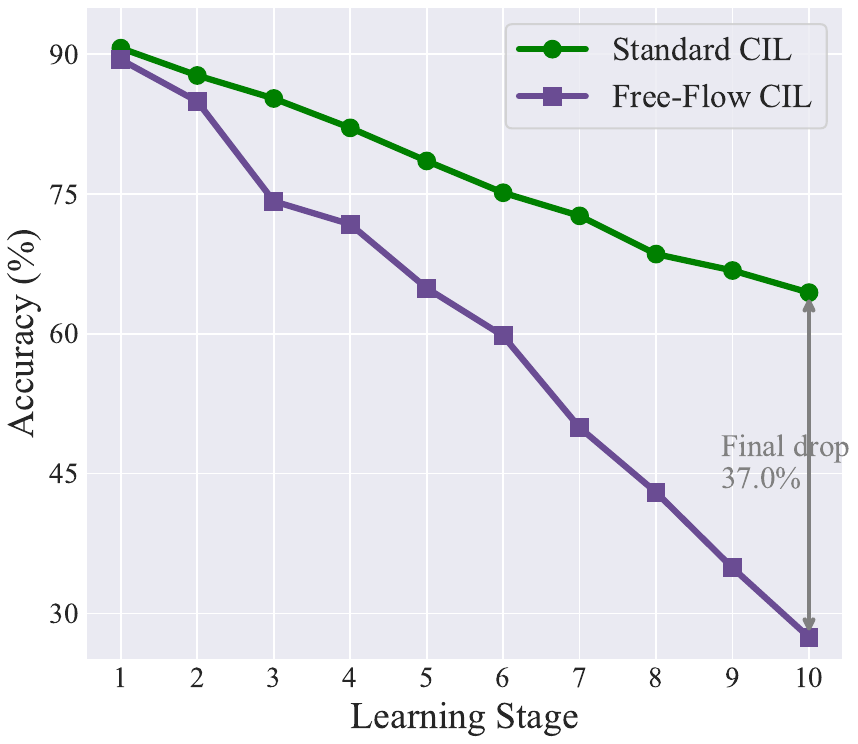}}  
  \caption{Illustration of FFCIL. (a) Unlike equal-size tasks, FFCIL allows variable per-step class increments. (b) Existing CIL methods could experience a substantial accuracy drop under FFCIL, even with the same total number of classes, learning stages, and global class permutation.}
  \label{fig:ffcildef}
\end{figure}

In practical class-arrival streams, updates are often triggered by the emergence of new categories rather than by a pre-defined balanced partition. As a result, the number of new classes introduced at each update can vary substantially, while the learner is still expected to incorporate each update immediately and classify all observed classes without task identities. We refer to this regime as Free-Flow Class-Incremental Learning (FFCIL). FFCIL preserves the standard CIL protocol, where classes arrive sequentially, and inference is performed over the accumulated label space, but relaxes the equal-increment assumption by allowing the class increment size to vary across learning stages.
This seemingly simple change creates a distinct source of difficulty. Variable class arrivals alter the class composition of training batches, make classifier calibration dependent on increment-size statistics, and weaken the comparability of representations learned under different class-arrival contexts. Consequently, as illustrated in Fig.~\ref{fig:ffcildef}, methods that perform well under equal-size increments can become unexpectedly brittle under FFCIL. This raises an important question: \textit{how can diverse CIL methods learn reliably under variable class arrivals?}

To address this challenge, we examine how variable class arrivals affect different components of existing CIL methods.
At the objective level, instance-wise loss aggregation assigns each class a coefficient determined by its frequency in the current mini-batch.
As the number of incoming classes changes, both the per-class contribution and the relative supervision between current and replay classes vary across steps.
At the classifier level, conventional weight alignment estimates the scale of new-class weights from the classes introduced at the current step.
Under FFCIL, the number of such weights varies substantially, making full alignment less reliable when only a few classes arrive.
For PTM-based methods, heterogeneous prediction mechanisms prevent uniform classifier-level alignment, while adaptation under different local class spaces can reduce feature comparability across increments. 
Based on these observations, we develop a general framework for robust learning under FFCIL.
Class-Wise Mean (CWM) replaces instance-wise aggregation with class-wise averaging to reduce step-dependent changes in class contribution.
Dynamic Intervention Weight Alignment (DIWA) adjusts the strength of new-class weight calibration according to the current increment size, avoiding excessive rescaling from unreliable statistics.
For PTM-based methods, Head-Agnostic Alignment (HA) performs a lightweight correction on intermediate features and uses previous-class feature statistics to retain supervision over the accumulated label space.
Together, these components improve the robustness of both conventional and PTM-based CIL methods under variable class arrivals.

Our contributions are summarized as follows:
\begin{itemize}
\item We formalize Free-Flow Class-Incremental Learning (FFCIL), which relaxes the equal-increment assumption and allows the number of newly arriving classes to vary across learning stages.

\item We propose a unified framework to improve robustness under FFCIL, addressing the above instabilities at the objective, classifier, and representation levels with Class-Wise Mean (CWM), Dynamic Intervention Weight Alignment (DIWA), and Head-agnostic Alignment (HA). 

\item We conduct extensive experiments on both conventional and PTM-based CIL methods across multiple datasets and class-arrival schedules. The results reveal consistent performance degradation under FFCIL and demonstrate that the proposed framework improves robustness.

\end{itemize}
\section{Related Work}

\subsection{Class-Incremental Learning}
A CIL model learns new classes over time and must classify samples at test time without task labels. Existing approaches can be broadly categorized into traditional methods without PTMs and recent PTM-based methods. Traditional CIL methods include replay-based methods~\cite{rebuffi2017icarl,wang2025enhancing,yang2024dynamic} that store or generate representative samples and reuse them in subsequent training, regularization and distillation-based methods~\cite{chen2022multi,bian2024make,fu2025enhancing} that preserve old knowledge by constraining parameter changes or matching previous model outputs and representations, and dynamic expansion methods~\cite{yan2021dynamically,zhou2022model,zheng2025task} that allocate additional parameters for new tasks or classes. In contrast, PTM-based methods \cite{wang2025tuna,gao2025knowledge} typically freeze a powerful pretrained backbone and perform continual adaptation by learning lightweight tuning modules or prediction heads, such as adapters \cite{sun2025mos} and LoRA \cite{he2025cllora} modules. Despite their differences, both traditional and PTM-based methods are commonly evaluated under relatively idealized protocols, where each step introduces a fixed or nearly fixed number of classes.

\subsection{Real-World Challenges in CIL}
Real-world deployments motivate CIL settings beyond the idealized benchmark. Few-Shot CIL (FSCIL) \cite{wang2023few,kim2025does} studies the case where each incremental step provides few labeled examples \cite{zhang2025few,cui2025few} for new classes. Class-Imbalanced CIL (also studied as long-tailed CIL) \cite{liu2022long,he2024gradient,xu2024defying} focuses on severe class imbalance, where head classes dominate, and tail classes are under-represented \cite{qi2025adaptive,lai2025tiny}. Task-Imbalanced Continual Learning (TICL)~\cite{hong2024dynamically} considers a setting where each task contains the same number of classes, but the number of training samples for some classes within certain tasks is abnormally high or low. These challenges highlight important practical \cite{dong2023no,raghavan2024online} factors such as data scarcity \cite{ma2025latest} and imbalance \cite{he2021tale}. Our Free-Flow Class-Incremental Learning targets a different realism gap: each incremental step may introduce an arbitrary number of new classes, and this number can vary drastically across consecutive updates.

\section{Preliminaries}
\subsection{Standard CIL Setup}
A CIL learner updates a classifier over a sequence of evolving datasets $D_1, D_2,$ 
$\ldots, D_t$~\cite{de2021continual}. Each $D_i$ is treated as an incremental task, which is a labeled dataset $D_i=\{(\mathbf{x}_j, y_j)\}_{j=1}^{n_i}$ with $n_i$ samples. In the standard CIL protocol,  the class set in $D_i$ is disjoint from those in all previous tasks and never reappears. At incremental step $t$, training is restricted to the current task $D_t$ together with a small exemplar memory drawn from previously seen classes~\cite{rebuffi2017icarl}. After step $t$, the learner has observed the union $D=\bigcup_{i=1}^{t} D_i$, and the label space expands to $\mathcal{Y}=Y_1 \cup Y_2 \cup \cdots \cup Y_t$, where $\mathcal{X}$ denotes the input space.
The learning objective is to train a predictor $f(\mathbf{x}):\mathcal{X}\rightarrow\mathcal{Y}$ that performs well on all classes learned so far. We evaluate performance on the cumulative test set $D^{test}=\bigcup_{i=1}^{t} D_i^{test}$ by minimizing the misclassification rate:
\begin{equation}
\label{eq:s3e1}
f^{*}(\mathbf{x})=\arg\min_{f\in\mathbb{H}} \mathbb{E}_{(\mathbf{x},y)\in D^{test}}\left[\mathbb{I}\big(f(\mathbf{x})\neq y\big)\right],
\end{equation}
where $\mathbb{H}$ is the hypothesis space and $\mathbb{I}(\cdot)$ is the indicator function.

\subsection{Problem Formulation of FFCIL}
Let $\mathcal{C}_t$ denote the label set associated with the incremental dataset $\mathcal{D}_t$. Most classical CIL benchmarks adopt a controlled, roughly balanced task split. For example, \emph{learning-from-scratch} \cite{rebuffi2017icarl} typically partitions the label space into tasks of equal size, i.e., $|\mathcal{C}_t| = |\mathcal{C}_{t-1}|$ for all $t$. In contrast, \emph{learning-from-half} \cite{douillard2020podnet} first learns a base session containing half of the classes and then learns the remaining classes in subsequent tasks with equal class counts.
However, such equal-task protocols only partially reflect real deployments. In practice, a trained model is often required to incorporate emerging concepts as soon as they appear in the stream, prompting immediate incremental updates driven by demand rather than a pre-defined balanced schedule; consequently, the class increment per step is irregular, i.e., $|\mathcal{C}_t|$ is not enforced to satisfy $|\mathcal{C}_t|=|\mathcal{C}_{t-1}|$.
Sometimes only $1$ to $2$ classes are introduced, while at other times a single update may bring in tens of classes. We formalize this regime as \emph{Free-Flow CIL (FFCIL)}, which allows arbitrarily varying and potentially bursty numbers of new classes in $D_t$. Specifically, the stream $\{\mathcal{D}_t\}_{t=1}^T$ is only required to satisfy:

\noindent\textbf{1) Free-flow.} Each step $t$ introduces a non-empty new class set $\mathcal{C}_t$ with highly variable size:
\begin{equation}
|\mathcal{C}_t|\ge 1,\ \ \big||\mathcal{C}_t|-|\mathcal{C}_{t-1}|\big|\ \text{is not fixed across steps}.
\end{equation}

\noindent\textbf{2) Non-repetition.} Previously observed classes do not reappear:
\begin{equation}
\mathcal{C}_t \cap \mathcal{C}_s = \varnothing,\quad \forall t\neq s.
\end{equation}

To instantiate FFCIL schedules, we first generate a feasible class-arrival spectrum under fixed total classes $C$ and steps $T$, with predefined minimum and maximum step sizes. 
We then temporally order the spectrum, such as by ascending, descending, or jumbled permutations, to form the final free-flow schedule. 
Detailed construction is provided in the supplementary material.

\begin{figure*}[!t]
  \centering
  \includegraphics[width=5.05in,keepaspectratio]{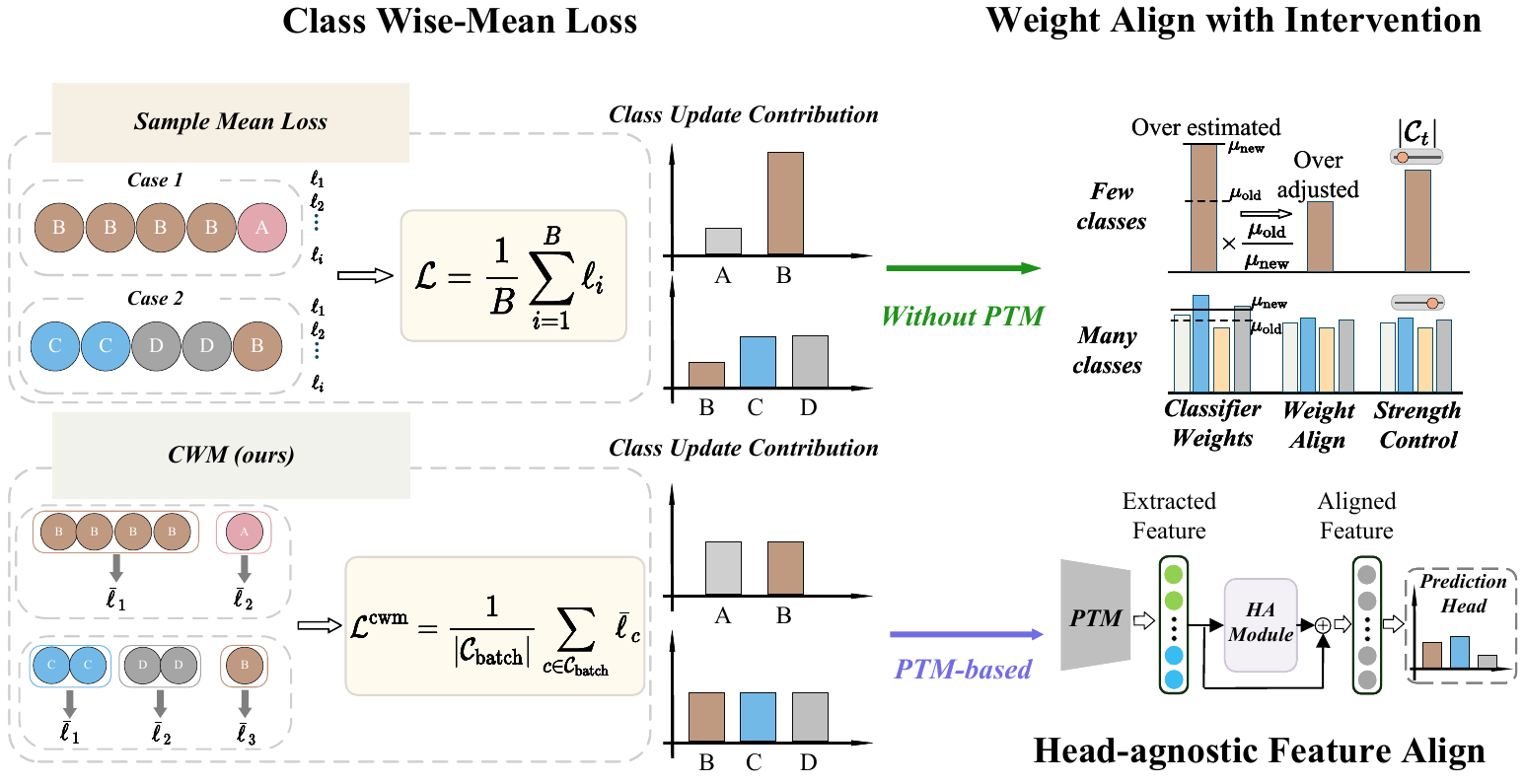}
  \caption{Overview of the proposed framework for FFCIL.
  Class-Wise Mean (CWM) replaces instance-wise loss aggregation with class-wise averaging to reduce the dependence of class contributions on step-specific batch composition.
  Dynamic Intervention Weight Alignment (DIWA) adjusts the strength of new-class weight calibration according to the number of classes introduced at the current step.
  Head-Agnostic Alignment (HA) performs a lightweight feature-level correction before the original prediction head and uses previous-class feature statistics to retain supervision over the accumulated label space.}
  \label{fig:ours}
\end{figure*}

\section{Learning under FFCIL}
To improve CIL robustness under variable class arrivals, we introduce three complementary components: Class-Wise Mean (CWM) for class-balanced supervision, Dynamic Intervention Weight Alignment (DIWA) for classifier calibration in non-PTM methods, and Head-Agnostic Alignment (HA) for feature alignment in PTM-based methods.
Non-PTM methods use CWM with DIWA, while PTM-based methods use CWM with HA, as illustrated in Fig.~\ref{fig:ours}.

\subsection{Class-Wise Mean Objective for FFCIL}
\label{newceloss}
Most objectives used in CIL are optimized by averaging per-sample losses within a mini-batch.
Taking cross-entropy (CE) as an example, for a mini-batch $b$ of size $B$, the conventional objective is
\begin{equation}
\label{eq:orgce}
\mathcal{L}_{\mathrm{CE}}
=
\frac{1}{B}
\sum_{i=1}^{B}
\ell_{\mathrm{CE}}\!\left(p_\theta(x_i), y_i\right).
\end{equation}
Let $n_c$ denote the number of samples of class $c$ in the batch, $b_c=\{i\in b\mid y_i=c\}$, and $\mathcal{C}_{\mathrm{batch}}$ denote the set of classes appearing in the batch.
Equation~\eqref{eq:orgce} can be equivalently written as
\begin{equation}
\label{eq:orgce2}
\mathcal{L}_{\mathrm{CE}}
=
\sum_{c\in\mathcal{C}_{\mathrm{batch}}}
\frac{n_c}{B}
\left(
\frac{1}{n_c}
\sum_{i\in b_c}
\ell_{\mathrm{CE}}\!\left(p_\theta(x_i),y_i\right)
\right).
\end{equation}
Thus, instance-wise averaging assigns class $c$ an effective coefficient $\pi_c=n_c/B$, which is determined by its frequency in the current mini-batch.
In CIL, mini-batches are constructed from the current increment and, when available, a limited set of replay samples.
When the number of newly arriving classes changes, the same sampling budget is distributed over a different class composition.
Consequently, both the per-class coefficients and the relative contribution between current and replay classes can vary across incremental steps.
This differs from conventional long-tailed learning, where the main concern is a fixed global class-frequency distribution; here, the class contribution changes with the incremental schedule.

To reduce this step-dependent frequency weighting, we propose the Class-Wise Mean (CWM) objective.
CWM first averages the loss within each class and then uniformly averages the resulting class-conditional losses:
\begin{equation}
\label{eq:cwm_ce}
\mathcal{L}^{\mathrm{cwm}}_{\mathrm{CE}}
=
\frac{1}{|\mathcal{C}_{\mathrm{batch}}|}
\sum_{c\in\mathcal{C}_{\mathrm{batch}}}
\left(
\frac{1}{n_c}
\sum_{i\in b_c}
\ell_{\mathrm{CE}}\!\left(p_\theta(x_i),y_i\right)
\right).
\end{equation}
Compared with Eq.~\eqref{eq:orgce2}, CWM replaces the empirical coefficient $n_c/B$ with $1/|\mathcal{C}_{\mathrm{batch}}|$.
Equivalently, if $\bar{\boldsymbol{g}}_c$ denotes the mean gradient of class $c$ in the batch, the two objectives yield
\begin{equation}
\label{eq:cwm_gradient}
\nabla_\theta\mathcal{L}_{\mathrm{CE}}
=
\sum_{c\in\mathcal{C}_{\mathrm{batch}}}
\frac{n_c}{B}\bar{\boldsymbol{g}}_c,
\nabla_\theta\mathcal{L}^{\mathrm{cwm}}_{\mathrm{CE}}
=
\frac{1}{|\mathcal{C}_{\mathrm{batch}}|}
\sum_{c\in\mathcal{C}_{\mathrm{batch}}}
\bar{\boldsymbol{g}}_c.
\end{equation}
CWM therefore removes the direct dependence of each class coefficient on its sample multiplicity in the batch, while retaining the original per-sample loss and optimization procedure.
Moreover, for prototype-based learners, CWM can be complemented with a center regularization term \cite{xu2025dual} to improve feature compactness and prototype estimation.
Let $\boldsymbol{h}_i$ denote the feature of sample $x_i$, and let $\boldsymbol{\mu}_{y_i}$ denote the feature center of class $y_i$.
The regularization term is
\begin{equation}
\label{eq:center_loss}
\mathcal{L}_{\mathrm{center}}
=
\frac{1}{B}
\sum_{i=1}^{B}
\left\|
\boldsymbol{h}_i-\boldsymbol{\mu}_{y_i}
\right\|_2^2.
\end{equation}
The resulting objective for prototype-based methods is
\begin{equation}
\label{eq:cwm_center_total}
\mathcal{L}_{\mathrm{total}}
=
\mathcal{L}^{\mathrm{cwm}}_{\mathrm{CE}}
+
\mathcal{L}_{\mathrm{center}}.
\end{equation}

Beyond the main classification loss, the same class-wise aggregation can be applied to other sample-wise objectives, such as the distillation loss \cite{rebuffi2017icarl} used in iCaRL-style methods and the auxiliary loss \cite{yan2021dynamically} $\mathcal{L}_{\mathrm{aux}}$ in DER.
\subsection{Dynamic Intervention Weight Alignment}

Conventional CIL methods commonly use a unified linear classifier over all observed classes.
After incremental training, the newly introduced classifier weights are optimized mainly with current-step data, whereas old-class weights are updated only through limited replay or distillation supervision.
This can produce a scale mismatch between old- and new-class weights and bias the resulting logits.
Weight alignment addresses this issue by rescaling the new-class weights according to the relative norms of the two groups.

Let $K$ be the number of previously learned classes and $C_t$ be the number of classes introduced at step $t$.
We denote the old- and new-class classifier weights by $\boldsymbol{W}_{\mathrm{old}}\in\mathbb{R}^{K\times d}$ and $\boldsymbol{W}_{\mathrm{new}}\in\mathbb{R}^{C_t\times d}$, respectively, and compute their average row norms as:
\begin{equation}
\label{eq:weight_norms}
\mu_{\mathrm{old}}
=
\frac{1}{K}
\sum_{c=1}^{K}
\|\boldsymbol{w}_c\|_2, \quad
\mu_{\mathrm{new}}
=
\frac{1}{C_t}
\sum_{c=K+1}^{K+C_t}
\|\boldsymbol{w}_c\|_2.
\end{equation}
Conventional weight alignment directly uses the ratio $\mu_{\mathrm{old}}/\mu_{\mathrm{new}}$ at every incremental step.
Under equal-size increments, $\mu_{\mathrm{new}}$ is estimated from a similar number of classifier weights across steps.
In FFCIL, however, this number is given by $C_t$ and can vary substantially.
When only a few classes arrive, the estimate can be sensitive to individual new-class weights, and full alignment may lead to excessive rescaling.
Under the standard finite-variance approximation for the new-class weight norms, the variance of their sample mean decreases with the increment size:
\begin{equation}
\label{eq:diwa_variance}
\operatorname{Var}\!\left(\mu_{\mathrm{new}}\right)
\approx
\frac{\sigma_w^2}{C_t},
\end{equation}
where $\sigma_w^2$ denotes the variance of new-class weight norms.
This motivates adapting the alignment strength to the number of classes observed at the current step.

We introduce Dynamic Intervention Weight Alignment (DIWA), which interpolates between retaining the original new-class weights and applying full weight alignment.
The intervention factor is defined as
\begin{equation}
\label{eq:diwa_eta}
\eta_t
=
1-(1-\eta_{\min})
\exp\!\left(
-\frac{C_t-1}{\tau}
\right),
\end{equation}
where $\eta_{\min}$ specifies the minimum intervention strength and $\tau$ controls how quickly the intervention saturates as $C_t$ increases.
The final scaling factor is
\begin{equation}
\label{eq:diwa_gamma}
\gamma_t
=
(1-\eta_t)
+
\eta_t
\frac{\mu_{\mathrm{old}}}{\mu_{\mathrm{new}}},
\quad
\boldsymbol{W}_{\mathrm{new}}
\leftarrow
\gamma_t\boldsymbol{W}_{\mathrm{new}}.
\end{equation}
Equation~\eqref{eq:diwa_gamma} reduces to retaining the original weights when $\eta_t=0$ and to conventional full weight alignment when $\eta_t=1$.
A small increment therefore leads to a conservative correction, whereas a larger increment permits stronger alignment because the new-class statistic is estimated from more classifier weights.
DIWA is applied once after each incremental training stage and introduces neither additional learnable parameters nor additional training iterations.

\subsection{Head-Agnostic Alignment for PTM-Based FFCIL}

DIWA is designed for conventional CIL methods with a unified linear classifier.
PTM-based CIL methods, however, employ heterogeneous prediction mechanisms, including prototype-based, analytic, and linear classifiers, so classifier-level weight alignment cannot be applied uniformly.
Moreover, their adaptation modules or prediction components are optimized under the local classification space of each increment.
A small increment involves relatively weak class competition, whereas a large increment requires separating many new classes simultaneously.
When the resulting representations are compared over the accumulated label space, this difference can lead to inconsistent feature distributions across increments.

To address this issue, we introduce Head-Agnostic Alignment (HA), which performs feature-level correction before method-specific prediction.
Let $\boldsymbol{h}=F_t(\boldsymbol{x})\in\mathbb{R}^{d}$ denote the feature produced by a PTM-based learner at step $t$.
HA applies a lightweight residual transformation:
\begin{equation}
\label{eq:ha_mapping}
\hat{\boldsymbol{h}}
=
\Phi_{\mathrm{HA}}(\boldsymbol{h})
=
\boldsymbol{h}
+
\boldsymbol{W}_{\mathrm{up}}
\operatorname{GELU}\!\left(
\boldsymbol{W}_{\mathrm{down}}
\operatorname{LN}(\boldsymbol{h})
\right),
\end{equation}
where $\operatorname{LN}(\cdot)$ denotes layer normalization.
Because HA operates on the intermediate feature rather than on classifier parameters, it is compatible with prototype-based, analytic, and linear prediction rules.
After the original CIL method finishes training at step $t$, all of its parameters are frozen and only HA is optimized on top of the fixed feature interface.

Training HA only with current-step features would bias the correction toward the current increment.
We therefore retain the class-wise feature mean $\boldsymbol{\mu}_c$ and standard deviation $\boldsymbol{\sigma}_c$ for each observed class $c$, where $\boldsymbol{\mu}_c,\boldsymbol{\sigma}_c\in\mathbb{R}^{d}$.
For each previous class $c\in\mathcal{Y}_{1:t-1}$, we generate $M$ surrogate features as:
\begin{equation}
\label{eq:ha_surrogate}
\tilde{\boldsymbol{h}}_c^m
=
\boldsymbol{\mu}_c
+
\boldsymbol{\epsilon}_c^m\odot\boldsymbol{\sigma}_c,
\quad
\boldsymbol{\epsilon}_c^m
\sim
\mathcal{N}(\boldsymbol{0},\boldsymbol{I}),
\quad
m=1,\ldots,M,
\end{equation}
where $\odot$ denotes element-wise multiplication.
These generated features provide previous-class supervision without storing original training samples.
We set $M=20$ in all experiments.

Let $\mathcal{B}_t$ denote the union of current-step features and generated surrogate features of previous classes.
The HA objective is:
\begin{equation}
\label{equ:ha}
\mathcal{L}_{\mathrm{HA}}
=
\frac{1}{|\mathcal{B}_t|}
\sum_{(\boldsymbol{h},y)\in\mathcal{B}_t}
\left[
\mathcal{L}_{\mathrm{cls}}\!\left(
\Phi_{\mathrm{HA}}(\boldsymbol{h}),y
\right)
+
\left\|
\Phi_{\mathrm{HA}}(\boldsymbol{h})-\boldsymbol{h}
\right\|_2^2
\right].
\end{equation}
The classification term preserves discriminability under the original prediction rule, while the identity term restricts HA from excessively altering the pretrained representation.
By optimizing the same feature-level correction with both current and previous classes, HA could reduce the bias toward the current arrival context and improve feature comparability over the accumulated label space.
The classification term $\mathcal{L}_{\mathrm{cls}}$ is instantiated according to the original prediction rule and can use the CWM objective.

\section{Experiments}
This section conducts extensive experiments. Sec.~5.2 investigates the performance of common CIL baselines on the FFCIL benchmark and validates the effectiveness of our framework. Sec.~5.3 presents ablation studies of each proposed component. Sec.~5.4 studies the impact of different step-size schedules on FFCIL. 
\subsection{Experimental Setup}
\textbf{Baselines.}  
We evaluate multiple baselines spanning both traditional and PTM-based CIL paradigms. For traditional CIL methods, we select iCaRL~\cite{rebuffi2017icarl}, DER~\cite{yan2021dynamically}, MEMO~\cite{zhou2022model}, and TagFex~\cite{zheng2025task}, covering representative replay-based and dynamic-expansion approaches. For PTM-based CIL methods, we choose several recent high-performing methods, including MoAL~\cite{gao2025knowledge}, MOS~\cite{sun2025mos}, TUNA~\cite{wang2025tuna}, and CL-LoRA~\cite{he2025cllora}.

\noindent\textbf{Implementation Details.} 
All methods are implemented in PyTorch. The baseline implementations and default hyperparameter settings mainly follow PyCIL~\cite{zhou2023pycil}, PILOT~\cite{zhou2024continual}, and~\cite{gao2025knowledge,he2025cllora}. 
For traditional baselines, iCaRL, DER, MEMO, and TagFex already include the fixed weight aligning strategy~\cite{zhao2020maintaining} in their original implementations, while our DIWA replaces this fixed WA.

\noindent\textbf{Evaluation Metrics.} 
Following the benchmark protocol~\cite{rebuffi2017icarl}, we use $A_t$ to denote the accuracy at stage $t$ on the test set containing all known classes after training with $D_1, D_2, \cdots, D_t$. $ A_T$ denotes the final-stage accuracy, evaluated on the test set that covers all learned classes. We also report the average accuracy $\overline{A}=\frac{1}{T}\sum_{t=1}^{T}A_t$, which summarizes the overall performance across the entire incremental learning process.

\subsection{Free-Flow Benchmark Comparsion}

\begin{table*}[!t]
\centering

\definecolor{GoodGreen}{RGB}{34,139,34}
\definecolor{BadRed}{RGB}{178,34,34}

\newcommand{\mainacc}[1]{%
    \scalebox{1.18}{$#1$}%
}

\newcommand{\eqaux}[1]{%
    \scalebox{0.74}{$\pm #1$}%
}

\newcommand{\orgaux}[3]{%
    \scalebox{0.74}{%
        $\pm #1\;\textcolor{BadRed}{#2#3}$%
    }%
}

\newcommand{\oursaux}[3]{%
    \scalebox{0.74}{%
        $\pm #1\;\textcolor{GoodGreen}{#2#3}$%
    }%
}

\newcommand{\eqresult}[2]{%
    \shortstack[c]{%
        \mainacc{#1}\\[-0.65mm]
        \eqaux{#2}%
    }%
}

\newcommand{\orgresult}[4]{%
    \shortstack[c]{%
        \mainacc{#1}\\[-0.65mm]
        \orgaux{#2}{#3}{#4}%
    }%
}

\newcommand{\oursresult}[4]{%
    \shortstack[c]{%
        \mainacc{#1}\\[-0.65mm]
        \oursaux{#2}{#3}{#4}%
    }%
}

\newcommand{\NA}{--}

\setlength{\tabcolsep}{1.0pt}
\renewcommand{\arraystretch}{1.10}

\resizebox{0.94\textwidth}{!}{
\begin{tabular}{@{}l*{12}{c}@{}}
\toprule

\multirow{3}{*}{\textbf{Non-PTM}}
& \multicolumn{6}{c}{\textbf{CIFAR-100}}
& \multicolumn{6}{c}{\textbf{ImageNet-100}} \\
\cmidrule(lr){2-7}
\cmidrule(lr){8-13}

& \multicolumn{2}{c}{\textbf{Equ.T}}
& \multicolumn{2}{c}{\textbf{FF. org}}
& \multicolumn{2}{c}{\textbf{FF. ours}}
& \multicolumn{2}{c}{\textbf{Equ.T}}
& \multicolumn{2}{c}{\textbf{FF. org}}
& \multicolumn{2}{c}{\textbf{FF. ours}} \\
\cmidrule(lr){2-3}
\cmidrule(lr){4-5}
\cmidrule(lr){6-7}
\cmidrule(lr){8-9}
\cmidrule(lr){10-11}
\cmidrule(lr){12-13}

& $\overline{A}$ & $A_T$
& $\overline{A}$ & $A_T$
& $\overline{A}$ & $A_T$
& $\overline{A}$ & $A_T$
& $\overline{A}$ & $A_T$
& $\overline{A}$ & $A_T$ \\
\midrule

iCaRL
& \eqresult{70.25}{0.12}
& \eqresult{49.80}{0.22}
& \orgresult{55.90}{1.50}{\downarrow}{14.35}
& \orgresult{42.96}{1.50}{\downarrow}{6.84}
& \oursresult{62.60}{1.30}{\uparrow}{6.70}
& \oursresult{49.24}{1.40}{\uparrow}{6.28}
& \eqresult{69.86}{0.12}
& \eqresult{51.54}{0.24}
& \orgresult{54.48}{1.90}{\downarrow}{15.38}
& \orgresult{36.44}{2.50}{\downarrow}{15.10}
& \oursresult{60.73}{1.70}{\uparrow}{6.25}
& \oursresult{43.10}{2.20}{\uparrow}{6.66} \\

DER
& \eqresult{74.44}{0.06}
& \eqresult{60.55}{0.11}
& \orgresult{71.87}{1.20}{\downarrow}{2.57}
& \orgresult{58.21}{1.60}{\downarrow}{2.34}
& \oursresult{72.28}{0.80}{\uparrow}{0.41}
& \oursresult{60.21}{1.00}{\uparrow}{2.00}
& \eqresult{77.62}{0.08}
& \eqresult{70.98}{0.16}
& \orgresult{76.40}{1.30}{\downarrow}{1.22}
& \orgresult{66.30}{1.70}{\downarrow}{4.68}
& \oursresult{77.91}{0.90}{\uparrow}{1.51}
& \oursresult{69.26}{1.10}{\uparrow}{2.96} \\

MEMO
& \eqresult{73.75}{0.16}
& \eqresult{59.07}{0.30}
& \orgresult{70.84}{1.10}{\downarrow}{2.91}
& \orgresult{56.52}{1.40}{\downarrow}{2.55}
& \oursresult{71.42}{0.80}{\uparrow}{0.58}
& \oursresult{58.34}{1.00}{\uparrow}{1.82}
& \eqresult{74.47}{0.16}
& \eqresult{63.34}{0.34}
& \orgresult{72.05}{1.50}{\downarrow}{2.42}
& \orgresult{57.90}{2.00}{\downarrow}{5.44}
& \oursresult{72.48}{1.00}{\uparrow}{0.43}
& \oursresult{61.70}{1.40}{\uparrow}{3.80} \\

TagFex
& \eqresult{82.98}{0.06}
& \eqresult{71.91}{0.13}
& \orgresult{80.04}{1.00}{\downarrow}{2.94}
& \orgresult{70.02}{1.20}{\downarrow}{1.89}
& \oursresult{80.38}{0.70}{\uparrow}{0.34}
& \oursresult{71.56}{0.90}{\uparrow}{1.54}
& \eqresult{77.05}{0.11}
& \eqresult{70.14}{0.19}
& \orgresult{74.88}{1.30}{\downarrow}{2.17}
& \orgresult{67.58}{1.60}{\downarrow}{2.56}
& \oursresult{75.50}{0.80}{\uparrow}{0.62}
& \oursresult{68.74}{1.10}{\uparrow}{1.16} \\

\midrule

\textbf{PTM-based}
& \multicolumn{6}{c}{\textbf{CARS}}
& \multicolumn{6}{c}{\textbf{ImageNet-A}} \\
\midrule

MOS
& \eqresult{79.42}{1.13}
& \eqresult{70.66}{1.28}
& \orgresult{76.12}{1.70}{\downarrow}{3.30}
& \orgresult{67.04}{2.00}{\downarrow}{3.62}
& \oursresult{78.19}{1.30}{\uparrow}{2.07}
& \oursresult{69.90}{1.50}{\uparrow}{2.86}
& \eqresult{65.71}{1.10}
& \eqresult{55.71}{0.90}
& \orgresult{62.82}{1.40}{\downarrow}{2.89}
& \orgresult{53.79}{1.60}{\downarrow}{1.92}
& \oursresult{64.43}{1.10}{\uparrow}{1.61}
& \oursresult{55.37}{1.10}{\uparrow}{1.58} \\

MoAL
& \eqresult{76.87}{1.20}
& \eqresult{62.42}{1.30}
& \orgresult{56.08}{6.00}{\downarrow}{20.79}
& \orgresult{25.41}{8.00}{\downarrow}{37.01}
& \oursresult{66.51}{3.50}{\uparrow}{10.43}
& \oursresult{53.58}{4.70}{\uparrow}{28.17}
& \eqresult{72.87}{1.00}
& \eqresult{63.53}{0.80}
& \orgresult{69.76}{2.00}{\downarrow}{3.11}
& \orgresult{57.34}{2.40}{\downarrow}{6.19}
& \oursresult{70.15}{1.20}{\uparrow}{0.39}
& \oursresult{60.76}{1.30}{\uparrow}{3.42} \\

CL-LoRA
& \eqresult{73.58}{1.10}
& \eqresult{61.26}{1.20}
& \orgresult{49.45}{8.00}{\downarrow}{24.13}
& \orgresult{21.17}{10.40}{\downarrow}{40.09}
& \oursresult{53.04}{5.20}{\uparrow}{3.59}
& \oursresult{35.63}{6.70}{\uparrow}{14.46}
& \eqresult{69.14}{2.23}
& \eqresult{58.26}{0.63}
& \orgresult{62.16}{3.30}{\downarrow}{6.98}
& \orgresult{50.95}{3.20}{\downarrow}{7.31}
& \oursresult{63.93}{2.10}{\uparrow}{1.77}
& \oursresult{54.71}{2.00}{\uparrow}{3.76} \\

TUNA
& \eqresult{83.85}{0.39}
& \eqresult{76.71}{0.35}
& \orgresult{77.52}{1.60}{\downarrow}{6.33}
& \orgresult{67.77}{2.10}{\downarrow}{8.94}
& \oursresult{82.05}{1.70}{\uparrow}{4.53}
& \oursresult{75.37}{2.20}{\uparrow}{7.60}
& \eqresult{73.25}{0.70}
& \eqresult{64.19}{1.00}
& \orgresult{69.77}{1.70}{\downarrow}{3.48}
& \orgresult{58.66}{2.00}{\downarrow}{5.53}
& \oursresult{72.03}{1.60}{\uparrow}{2.26}
& \oursresult{64.08}{1.80}{\uparrow}{5.42} \\

\bottomrule
\end{tabular}
}
\caption{$\overline{A}$ and $A_T$ under Equ.T, FF.org, and FF.ours. Results are reported as mean $\pm$ standard deviation over five seeds for each applicable randomness: the global class permutation for Equ.T, and the global class permutation, class-arrival spectrum, and temporal schedule for FFCIL. Arrows denote the drop from Equ.T to FF.org and the recovery from FF.org to FF.ours.}
\label{tab:bc}
\end{table*}

In this subsection, we evaluate representative baselines under both the standard CIL protocol and the FFCIL protocol. We build an FFCIL protocol on CIFAR-100~\cite{krizhevsky2009learning} dataset in which the number of newly arriving classes ranges from 1 to 15 in incremental steps, and compare three settings: standard CIL with equal class splits (Equ.T), FFCIL using the original baseline methods (FF.org), and FFCIL equipped with our framework (FF.ours). For a fair comparison, Equ.T and FFCIL use the same total number of learning steps. Since CIFAR-100 is a relatively small-scale dataset, we further conduct experiments on the larger-scale ImageNet-100 \cite{deng2009imagenet} dataset. Moreover, to isolate the effect of free-flow class arrival from class-order randomness, Equ.T and FFCIL are constructed from the same global class permutation. The only difference is the partition boundary, i.e., the number of new classes introduced at each step. The results are summarized in Table~\ref{tab:bc}. 
Across datasets of different scales, even with the same total number of learning steps, replacing equal class splits with free-flow arrivals causes clear accuracy drops for both distillation and expansion-based methods, with degradation reaching up to 15\%. This confirms that free-flow class arrival itself can substantially degrade CIL performance.
In contrast, our framework consistently improves performance across different baselines.

We similarly evaluate PTM-based CIL methods on two commonly used benchmarks, CARS~ \cite{kramberger2020lsun} and ImageNet-A~\cite{hendrycks2021natural}. As shown in Table~\ref{tab:bc}, all PTM-based methods suffer from accuracy degradation under FFCIL, whereas our framework yields consistent improvements. Notably, although TUNA achieves much higher accuracy than MOS under the equal-step setting, it becomes clearly inferior to MOS under the free-flow setting. A possible reason is that MOS trains task-wise adapters and performs model-agnostic aggregation at inference time. In contrast, TUNA, MoAL, and CL-LoRA introduce task-shared components to fuse, select, or modulate knowledge across increments. Such components implicitly require features from different increments to remain comparable under a common representation interface. Under FFCIL, however, free-flow class arrivals make this interface less stable, leading to more pronounced degradation. 

\subsection{Ablation Study}

\newcolumntype{C}{>{\centering\arraybackslash}X}
\begin{table}[!t]
    \centering
    \setlength{\tabcolsep}{3pt}
    \renewcommand{\arraystretch}{1.08}

    \begin{tabularx}{0.9\columnwidth}{@{}ccCCCC@{}}
    \toprule
    \multirow{2}{*}{CWM} 
    & \multirow{2}{*}{DIWA}
    & \multicolumn{2}{c}{iCaRL}
    & \multicolumn{2}{c}{DER} \\
    \cmidrule(lr){3-4} \cmidrule(lr){5-6}
    &
    & $\overline{A}$ & $A_T$
    & $\overline{A}$ & $A_T$ \\
    \midrule
    $\times$ & $\times$ 
    & 55.90 & 42.96 
    & 71.87 & 58.21 \\

    \checkmark & $\times$ 
    & 58.88 & 45.29
    & 72.17 & 60.01 \\

    \checkmark & \checkmark 
    & \textbf{62.60} & \textbf{49.24}
    & \textbf{72.28} & \textbf{60.21} \\

    \midrule
    \multirow{2}{*}{CWM} 
    & \multirow{2}{*}{HA}
    & \multicolumn{2}{c}{MOS}
    & \multicolumn{2}{c}{TUNA} \\
    \cmidrule(lr){3-4} \cmidrule(lr){5-6}
    &
    & $\overline{A}$ & $A_T$
    & $\overline{A}$ & $A_T$ \\
    \midrule
    $\times$ & $\times$ 
    &  76.12 & 67.04 
    & 77.52 & 67.77 \\

    \checkmark & $\times$ 
    &  77.03 &  68.05
    & 79.23 & 70.83 \\

    \checkmark & \checkmark 
    & \textbf{78.19} & \textbf{69.90}
    & \textbf{82.05} & \textbf{75.37} \\
    \midrule
    \multirow{2}{*}{HA-$\mathcal{L}_{\text{idn}}$} 
    & \multirow{2}{*}{HA-SF}
    & \multicolumn{2}{c}{MOS}
    & \multicolumn{2}{c}{TUNA} \\
    \cmidrule(lr){3-4} \cmidrule(lr){5-6}
    &
    & $\overline{A}$ & $A_T$
    & $\overline{A}$ & $A_T$ \\
    \midrule
    \checkmark & $\times$
    & 77.40 & 68.34
    & 79.85 & 71.13 \\
    $\times$ & \checkmark
    & 77.61 & 69.16
    & 81.69 & 74.84 \\
    \bottomrule
    \end{tabularx}
    \caption{Ablation study of FFCIL components.}
    \label{tab:abl}
\end{table}

In this section, we conduct ablation studies on the proposed components, as shown in Table~\ref{tab:abl}. The results demonstrate that CWM consistently improves all CIL methods. For conventional CIL methods, the original baselines already adopt weight alignment, while replacing it with DIWA further improves performance. This indicates that adaptively adjusting the alignment strength according to the current step size is necessary under FFCIL. For PTM-based methods, HA provides clear performance gains. Moreover, the internal ablations of HA verify the usefulness of its key designs. HA-$\mathcal{L}_{\mathrm{idn}}$ confirms the importance of the identity constraint in preventing over-correction of PTM features. HA-SF shows that statistical surrogate features are crucial. Without them, HA is trained only within the label space on the current step, preventing it from forming a common feature-level correction over accumulated classes.

\begin{figure}[!t]
  \centering
  \subfloat[CWM on training]{
      \includegraphics[width=1.5in,keepaspectratio]{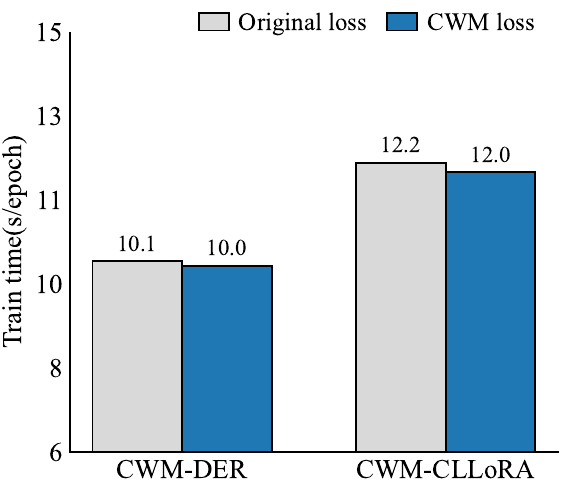}}
   \subfloat[HA on inference]{
      \includegraphics[width=1.5in,keepaspectratio]{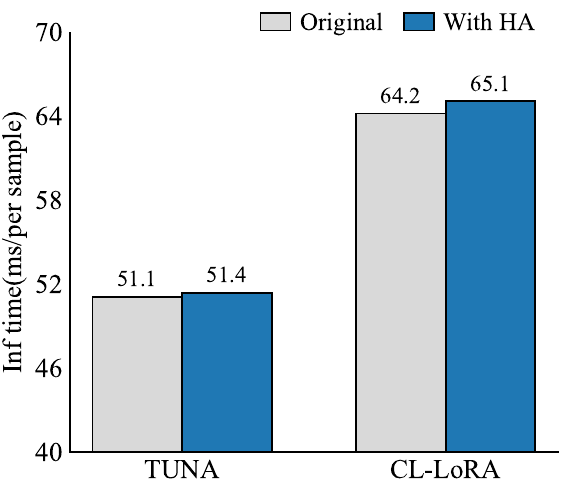}}   
      
  \caption{Computational impact study of the proposed CWM loss and HA.}
  \label{fig:abl2}
\end{figure}

In addition, we examine the computational impact of the components on training and inference. Fig.~\ref{fig:abl2} reports the training-time overhead of CWM and the inference-time overhead of HA. The results show that CWM can even slightly reduce the training time due to its simple class-wise aggregation. HA only adds a lightweight module before prediction, leading to a very small increase in inference time. These results indicate that the proposed framework improves robustness under FFCIL without imposing noticeable computational overhead.

\subsection{Impact of Step-Size Schedules in FFCIL}
\begin{figure}[!t]
  \centering
  \subfloat[non-PTM]{
      \includegraphics[width=1.55in,keepaspectratio]{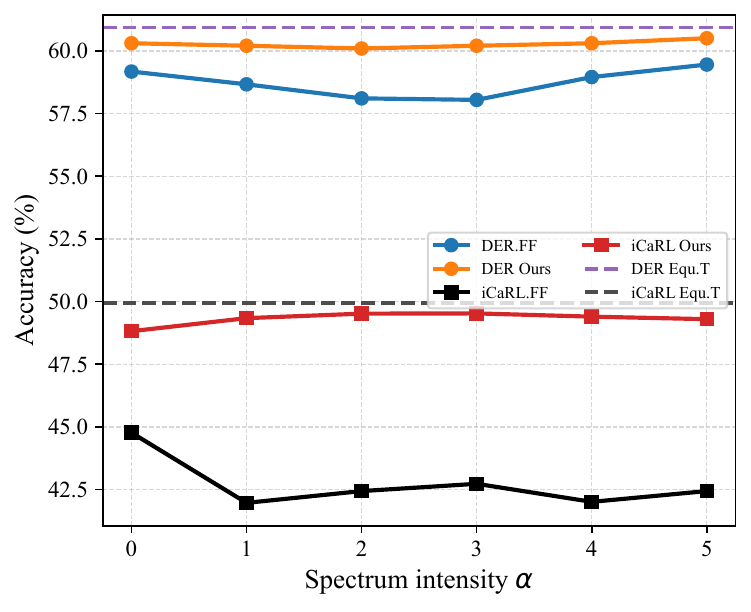}}
   \subfloat[PTM-based]{
      \includegraphics[width=1.55in,keepaspectratio]{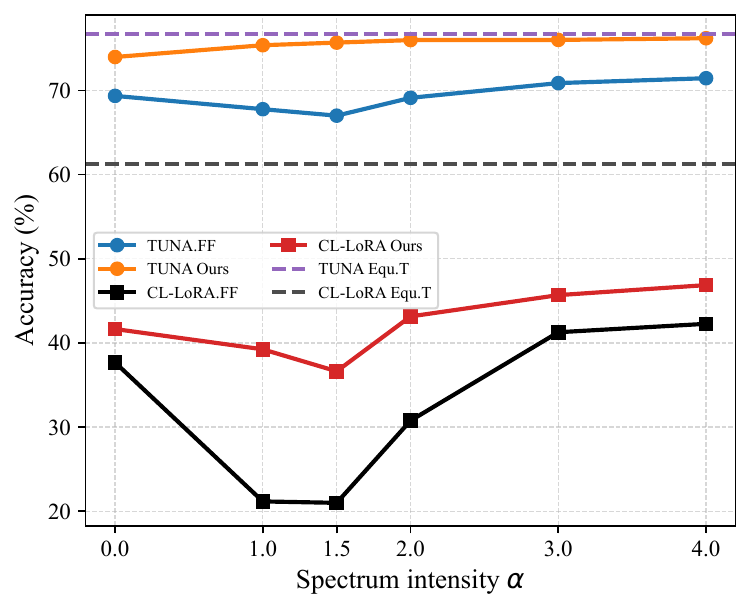}}  

  \caption{Impact of FFCIL class-arrival spectra under different values of the spectrum intensity parameter \(\alpha\).}
  \label{fig:stepspe}
\end{figure}

We first study how the class-arrival spectrum affects performance. As described in the supplementary material, we keep the global class permutation and the temporal ordering strategy fixed, and vary the spectrum intensity by changing \(\alpha\). Experiments are conducted on CIFAR-100 with DER and iCaRL, and on CARS with CL-LoRA and TUNA, as shown in Fig.~\ref{fig:stepspe}. The results show that performance degradation appears across different spectrum intensities, indicating that FFCIL is not an artifact of a particular hand-crafted schedule, but a general challenge caused by irregular class arrivals. Moreover, excessively increasing \(\alpha\) does not necessarily make the stream more difficult, because the class budget becomes increasingly concentrated near the boundary step sizes, reducing the diversity of intermediate increments. In contrast, a moderate \(\alpha\) can produce more varied and uneven middle-sized steps, making the learning process more irregular. Our method consistently improves the baselines under different spectrum intensities.

\begin{figure}[!t]
  \centering
  \subfloat[iCaRL]{
      \includegraphics[width=1.5in,keepaspectratio]{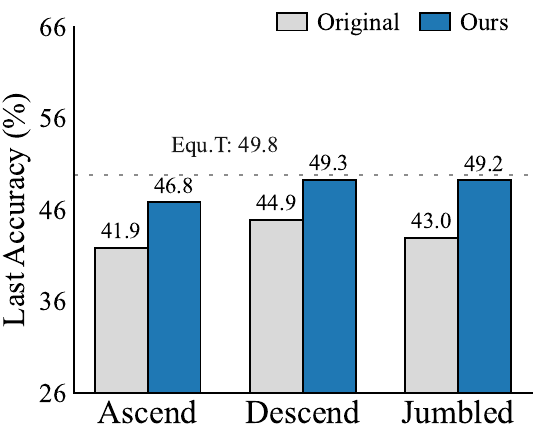}}
   \subfloat[DER]{
      \includegraphics[width=1.5in,keepaspectratio]{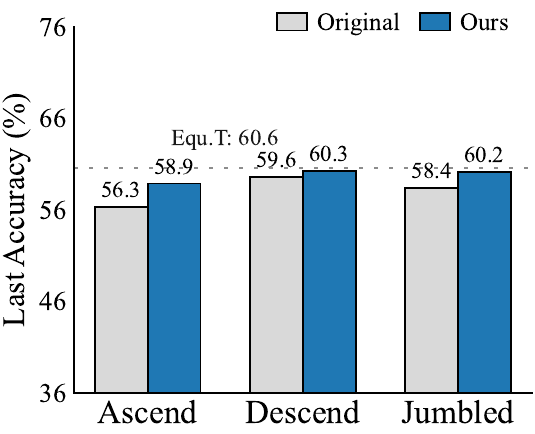}}   
      
  \subfloat[CL-LoRA]{
      \includegraphics[width=1.5in,keepaspectratio]{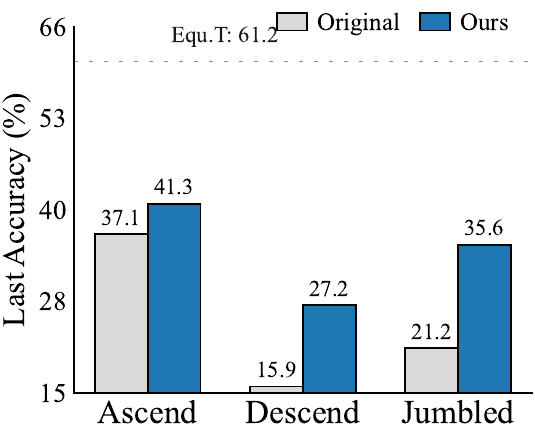}}
   \subfloat[TUNA]{
      \includegraphics[width=1.5in,keepaspectratio]{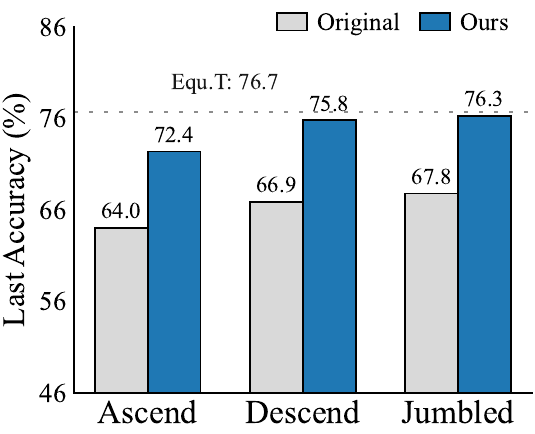}}   
  \caption{Impact of FFCIL step schedules under ascending, descending, and jumbled schedules.}
  \label{fig:stepsche}
\end{figure}

We further study how different FFCIL schedules affect performance. As described in the supplementary material, we compare three schedules with the same class-arrival spectrum and permutation, differing only in the temporal ordering, including the main \textit{jumbled} setting as well as the \textit{ascending} and \textit{descending} schedules, as shown in Fig.~\ref{fig:stepsche}. The results show that FFCIL reflects a general phenomenon caused by irregular class arrivals, rather than an artifact of a carefully designed schedule. All temporal orderings lead to clear performance changes, and different methods exhibit distinct sensitivities, e.g., CL-LoRA degrades severely under the descending schedule, while many methods are more affected by the ascending schedule. Our method consistently improves performance across these schedules.

\section{Conclusion}
We introduced Free-Flow Class-Incremental Learning (FFCIL), a CIL setting that relaxes the commonly used equal-increment assumption and studies learning under variable class arrivals.
Our analysis shows that changing the number of incoming classes affects class contributions in incremental objectives, the reliability of new-class weight statistics, and the consistency of PTM features learned across increments.
To address these issues, CWM replaces instance-wise aggregation with class-wise averaging, DIWA adapts new-class weight alignment to the current increment size, and HA performs head-agnostic feature correction using current and previous-class supervision.
Experiments across conventional and PTM-based CIL methods demonstrate that variable class arrivals cause consistent performance degradation, while the proposed framework improves robustness across different datasets and arrival schedules with limited computational overhead.
We hope FFCIL encourages future CIL research to consider class-arrival variability when designing and evaluating incremental learners.
\bibliography{main}

\end{document}